\documentclass[11pt]{article}
\usepackage[final]{acl}    % "final" hides the review boxes and submission-only notes
\usepackage{fontspec}
\newfontfamily\nepalifont{NewCM10Devanagari}[
  Extension   = .otf,
  UprightFont = *-Regular,
  BoldFont    = *-Bold,
  Script      = Devanagari
]
\usepackage{latexsym}
\usepackage{microtype}
\usepackage{graphicx}
\usepackage{booktabs}
\usepackage{multirow}
\usepackage{xcolor}
\usepackage{amsmath}
\usepackage{amssymb}
\usepackage{enumitem}
\usepackage{url}
\usepackage{xurl}        % allow URLs to break at any character
\usepackage{hyperref}
\hypersetup{colorlinks=true,linkcolor=black,citecolor=black,urlcolor=blue}

\title{Comparative Analysis of Multilingual Pre-trained Models for Nepali Automatic Speech Recognition}

\author{Suman Paudel \\
  School of Mathematical Sciences \\
  Institute of Science and Technology \\
  Tribhuvan University \\
  Kathmandu, Nepal \\
  \texttt{dastonsuman1997@gmail.com} \\
  \And
  Asst.\ Prof.\ Sarbin Sayami \\
  Central Department of Computer Science \\
  and Information Technology \\
  Tribhuvan University \\
  Kathmandu, Nepal \\
}

\begin{document}
\maketitle

\begin{abstract}
Multilingual pretrained models nominally support Nepali, yet no controlled benchmark has compared them under a single fine-tuning protocol. We fine-tune six pretrained models (XLSR-53, IndicWav2Vec, MMS-1B, Whisper-Medium, Whisper-Large-v3-Turbo, and Conformer-Hi) spanning CTC self-supervised, autoregressive encoder--decoder, and hybrid Conformer-CTC architectures, on the OpenSLR SLR54 Nepali corpus ($\sim$165~hours) using identical preprocessing, splits, optimizer, and family-matched learning-rate schedules. We evaluate Word Error Rate (WER), Character Error Rate (CER), and Real-Time Factor (RTF) on three independent test sets (OpenSLR, FLEURS, Common Voice). Whisper-Large-v3-Turbo (14.76\% WER) and IndicWav2Vec (14.89\% WER) tie at the top despite a 9$\times$ parameter gap and 40$\times$ pretraining-data gap, providing direct empirical evidence that \emph{language-family proximity in pretraining can substitute for raw scale} for in-domain Nepali. CTC decoders run up to 29$\times$ faster than autoregressive Whisper at the same accuracy, flipping the practical deployment preference toward CTC under any latency budget. Massively multilingual pretraining (MMS-1B) yields the smallest out-of-domain degradation on FLEURS (+12.55~pp), indicating that scale buys robustness rather than peak in-domain accuracy. The resulting benchmark provides the first standardized, multi-model, efficiency-aware reference numbers for Nepali ASR.
\end{abstract}

\section{Introduction}

Self-supervised and weakly supervised multilingual pretraining have driven Automatic Speech Recognition (ASR) on resource-rich languages to near-human accuracy \citep{baevski2020wav2vec,radford2023robust}. The gains have not, however, propagated uniformly to the world's $\sim$7{,}000 languages \citep{besacier2014automatic,pratap2023scaling}.

Nepali (\textit{ne}) is an Indo-Aryan language with $\sim$32 million native speakers, written in Devanagari script. It combines contrastive aspirated/unaspirated stops, conjunct-heavy orthography, agglutinative morphology, and free word order. The volume of openly licensed transcribed Nepali speech is approximately 165~hours \citep{kjartansson2018open}, an order of magnitude less than English. Several multilingual pretrained models nominally support Nepali, either through explicit pretraining inclusion (MMS-1B, IndicWav2Vec) or through zero-shot multilingual capability (Whisper). However, published Nepali ASR results have so far come from disjoint single-model studies that each evaluated one model on one dataset with different preprocessing pipelines. Cross-model comparison has therefore been impossible, and no defensible model-selection guidance exists for Nepali practitioners.

This paper closes that gap. We fine-tune six multilingual pretrained ASR models from three architectural families on the OpenSLR~SLR54 Nepali corpus under an identical preprocessing and training protocol, then evaluate them on three independent test sets along three orthogonal axes: accuracy (WER, CER), inference efficiency (RTF), and Nepali-specific error patterns.

\textbf{Contributions.} (i)~The first standardized multi-model multi-dataset benchmark for Nepali ASR. (ii)~Empirical isolation of pretraining \emph{proximity} from pretraining \emph{scale}, showing they trade off rather than compound. (iii)~First-of-kind RTF measurements for Nepali ASR, exposing a 29$\times$ CTC-vs-autoregressive speed gap. (iv)~Per-scenario deployment recommendations grounded in measurement. (v)~Public release of all six fine-tuned checkpoints and a per-utterance reference/prediction benchmark dataset on the Hugging Face Hub.

\section{Related Work}

\paragraph{Self-supervised speech pretraining.}
Wav2Vec~2.0 \citep{baevski2020wav2vec} learns speech representations by contrastive prediction over masked latents. Multilingual extensions include XLSR-53 \citep{conneau2020unsupervised} (53 languages), XLS-R \citep{babu2021xls} (128 languages, 436~k~hr), HuBERT \citep{hsu2021hubert}, and WavLM \citep{chen2022wavlm}. The Massively Multilingual Speech (MMS) project \citep{pratap2023scaling} extended this recipe to $>$1{,}100 languages including Nepali, releasing MMS-1B with per-language CTC heads.

\paragraph{Encoder-decoder and hybrid architectures.}
Whisper \citep{radford2023robust} is an autoregressive Transformer encoder--decoder trained on $\sim$680~k~hr of weakly supervised multilingual web audio. The Conformer architecture \citep{gulati2020conformer} interleaves multi-head self-attention with depthwise-separable convolution; pretrained Conformer-CTC checkpoints exist for several Indic languages.

\paragraph{Indic-family ASR.}
IndicWav2Vec \citep{javed2021towards} pretrains Wav2Vec~2.0 on $\sim$17~k~hr of speech across 40 Indic languages, formalizing the hypothesis that language-family-proximate pretraining can outperform broader multilingual pretraining for Indic targets. Vakyansh \citep{kumar2022vakyansh}, Vistaar \citep{bhogale2023vistaar}, and IndicVoices \citep{javed2024indicvoices} extend the ecosystem with toolkits, benchmarks, and diverse corpora.

\paragraph{Nepali ASR.}
\citet{ghimire2023active} fine-tuned MMS-1B on Nepali with active-learning-based data selection (CER 6.77\%). MMS and IndicWav2Vec papers include Nepali but report only Indic aggregates. No prior work compares multiple architecture families on Nepali under a single protocol with efficiency and error-pattern measurements.

\section{Method}
\label{sec:method}

\subsection{Datasets}
\textbf{OpenSLR~SLR54} \citep{kjartansson2018open}: $\sim$165~hr of read Nepali speech from 527 volunteer speakers, partitioned 80/10/10 into training, validation, and test with \emph{speaker-disjoint} splits. \textbf{FLEURS (ne\_np)} \citep{conneau2023fleurs}: $\sim$10~hr of curated read Nepali from FLoRes-101 prompts, used unmodified as a controlled out-of-domain test set. \textbf{Common Voice (ne-NP)} \citep{ardila2020common}: $\sim$5~hr of crowd-sourced validated Nepali audio with substantial microphone, noise, and accent variability; used as the most challenging out-of-domain test set.

\subsection{Models}
The six evaluated models span three architectural families (CTC self-supervised, autoregressive encoder--decoder, hybrid Conformer-CTC), five pretraining strategies (Indic SSL, broadly multilingual SSL, massively multilingual SSL, supervised multilingual, language-family-proximate supervised), and a parameter range of more than 30$\times$ (30.5~M to 965~M).

\begin{table}[t]
\centering
\footnotesize
\setlength{\tabcolsep}{4pt}
\resizebox{\columnwidth}{!}{%
\begin{tabular}{llcr}
\toprule
\textbf{Model} & \textbf{Architecture Family} & \textbf{Decoder} & \textbf{Params} \\
\midrule
XLSR-53        & CTC SSL (Wav2Vec~2.0)         & CTC      & 317~M \\
IndicWav2Vec   & CTC SSL (Wav2Vec~2.0)         & CTC      & 94.4~M \\
MMS-1B         & CTC SSL (Wav2Vec~2.0)         & CTC      & 965~M \\
Whisper-Medium & Encoder--Decoder              & Autoreg. & 769~M \\
Whisper-Turbo  & Encoder--Decoder              & Autoreg. & 809~M \\
Conformer-Hi   & Hybrid Conformer-CTC          & CTC      & 30.5~M \\
\bottomrule
\end{tabular}}
\caption{Six evaluated models spanning three architectural families. ``Whisper-Turbo'' is Whisper-Large-v3-Turbo. Conformer-Hi is the Hindi-pretrained NVIDIA NeMo Conformer-CTC checkpoint.}
\label{tab:models}
\end{table}

\subsection{Preprocessing}
A single pipeline is applied to every dataset and every model so that comparisons are controlled: (1)~audio resampled to 16~kHz; (2)~stereo converted to mono; (3)~transcriptions Unicode-normalised to NFC, punctuation stripped; (4)~utterances $<$0.5~s or $>$30~s removed; (5)~raw waveform inputs for the Wav2Vec~2.0 family, 80-dim log-Mel spectrograms (25~ms / 10~ms) for the Whisper and Conformer-CTC families.

\subsection{Training configuration}
All Wav2Vec~2.0-family models use AdamW with learning rate $3\times10^{-4}$, batch size 8, 500-step warmup, and cosine decay over 10 epochs; the convolutional feature encoder is frozen. Whisper models use AdamW with $1\times10^{-5}$, batch size 8, linear warmup and decay over 3--6 epochs (all parameters trainable). Conformer-Hi is fine-tuned in NVIDIA NeMo with AdamW at $1\times10^{-4}$, batch size 16, cosine annealing over 10 epochs. SpecAugment \citep{park2019specaugment} is applied to all models. Early stopping is triggered when validation WER stagnates for three consecutive evaluation rounds.

\subsection{Experimental protocol}
\label{sec:protocol}
The end-to-end evaluation harness and the inference scripts used to reproduce the numbers reported below are released openly.\footnote{Code: \href{https://github.com/p-sumann/nepali-asr-benchmark}{github.com/p-sumann/nepali-asr-benchmark}.} Three sequential phases. \textbf{Phase~1 (zero-shot)}: each model is evaluated on the three test sets without Nepali-specific fine-tuning, establishing an out-of-the-box baseline and justifying the cost of Phase~2. \textbf{Phase~2 (controlled fine-tuning)}: all six models are fine-tuned on the OpenSLR~SLR54 training split under the shared protocol above. \textbf{Phase~3 (multi-test evaluation, efficiency)}: fine-tuned models are evaluated on all three test sets and RTF is measured on a single NVIDIA L4 GPU with batch size 1.

\subsection{Evaluation metrics}
WER (Equation~\ref{eq:wer}) and CER (Equation~\ref{eq:cer}) are computed with \texttt{jiwer} after NFC normalisation on both references and hypotheses. RTF (Equation~\ref{eq:rtf}) is the ratio of processing time to audio duration. CER is particularly informative for Nepali, where word-boundary ambiguity inflates WER and where character-level fidelity is needed to capture Devanagari orthographic detail.

\begin{equation}
\mathrm{WER} = \frac{S+D+I}{N}\times 100\%
\label{eq:wer}
\end{equation}
\begin{equation}
\mathrm{CER} = \frac{S_c+D_c+I_c}{N_c}\times 100\%
\label{eq:cer}
\end{equation}
\begin{equation}
\mathrm{RTF} = \frac{T_\mathrm{processing}}{T_\mathrm{audio}}
\label{eq:rtf}
\end{equation}

\section{Results}
\label{sec:results}

\subsection{Zero-shot performance}
Zero-shot WER and CER are reported on the four models for which Nepali zero-shot evaluation is meaningful (Conformer-Hi and XLSR-53 lack a Nepali-aware decoding head). Only MMS-1B produced independently usable output, and only on FLEURS (31.75\% WER). Whisper models predominantly hallucinated Hindi or English. IndicWav2Vec WER $>200\%$ reflects an uninitialised Nepali CTC head rather than meaningful performance. These results justify Phase~2: fine-tuning is essential for every model.

\begin{table}[t]
\centering
\footnotesize
\setlength{\tabcolsep}{3pt}
\resizebox{\columnwidth}{!}{%
\begin{tabular}{lcccccc}
\toprule
& \multicolumn{2}{c}{\textbf{OpenSLR}} & \multicolumn{2}{c}{\textbf{FLEURS}} & \multicolumn{2}{c}{\textbf{Common V.}} \\
\cmidrule(lr){2-3}\cmidrule(lr){4-5}\cmidrule(lr){6-7}
\textbf{Model} & W & C & W & C & W & C \\
\midrule
MMS-1B          & \textbf{55.94} & \textbf{16.48} & \textbf{31.75} & \textbf{8.48}  & \textbf{59.38} & \textbf{21.05} \\
Whisper-Turbo   & 115.1 & 45.31 & 94.09 & 29.67 & 92.79 & 42.72 \\
Whisper-Medium  & 114.5 & 52.44 & 110.3 & 39.98 & 94.46 & 39.71 \\
IndicWav2Vec    & 360.6 & 47.75 & 366.5 & 51.87 & 260.7 & 45.76 \\
\bottomrule
\end{tabular}}
\caption{Zero-shot WER (W) / CER (C) in \%.}
\label{tab:zero-shot}
\end{table}

\subsection{Fine-tuned in-domain results}
\label{sec:in-domain}
After fine-tuning, three observations stand out from the best validation metrics. First, the top three models (Whisper-Turbo, IndicWav2Vec, Whisper-Medium) fall within roughly one percentage point of each other on validation WER despite spanning a 9$\times$ range in parameter count and despite using fundamentally different decoders (autoregressive vs.\ CTC). Second, IndicWav2Vec reaches its best WER with 94.4~M parameters in under six hours of training, whereas Whisper-Turbo requires 809~M parameters and approximately 34 hours. Third, Conformer-Hi (30.5~M, Hindi-only pretraining) outperforms both XLSR-53 (317~M) and MMS-1B (965~M), reinforcing that linguistic proximity in pretraining matters more than raw model capacity.

\begin{table}[t]
\centering
\small
\setlength{\tabcolsep}{4pt}
\begin{tabular}{lrrrr}
\toprule
\textbf{Model} & \textbf{Params} & \textbf{WER} & \textbf{CER} & \textbf{Dur.} \\
\midrule
Whisper-Turbo   & 809~M  & \textbf{14.27} & 3.35 & 33.9~h \\
IndicWav2Vec    & 94.4~M & 15.08 & \textbf{3.15} & 5.9~h  \\
Whisper-Medium  & 769~M  & 15.12 & 3.81 & 17.9~h \\
Conformer-Hi    & 30.5~M & 24.68 & 5.82 & \textbf{3.2~h}  \\
XLSR-53         & 317~M  & 26.73 & 6.04 & 10.0~h \\
MMS-1B          & 965~M  & 26.99 & 6.09 & 19.2~h \\
\bottomrule
\end{tabular}
\caption{Best validation metrics during fine-tuning (\%). Duration is wall-clock training time on a single NVIDIA L4.}
\label{tab:training}
\end{table}

\subsection{Multi-test-set benchmark}
\label{sec:benchmark}
On the fine-tuned benchmarks, Whisper-Turbo achieves the lowest in-domain WER (14.76\%), with IndicWav2Vec (14.89\%) and Whisper-Medium (15.57\%) within a single percentage point. Conformer-Hi, XLSR-53, and MMS-1B form a middle tier around 26--27\% WER. On FLEURS, Whisper-Medium leads narrowly (39.06\% WER), but MMS-1B is the only model with $<$11\% CER on FLEURS, reflecting acoustic-condition diversity from its $>$1{,}100-language pretraining. On Common Voice, every model degrades sharply; even the best model crosses 48\% WER, establishing crowd-sourced Nepali audio as the largest remaining open problem.

\begin{table}[t]
\centering
\footnotesize
\setlength{\tabcolsep}{3pt}
\resizebox{\columnwidth}{!}{%
\begin{tabular}{lcccccc}
\toprule
& \multicolumn{2}{c}{\textbf{OpenSLR}} & \multicolumn{2}{c}{\textbf{FLEURS}} & \multicolumn{2}{c}{\textbf{Common V.}} \\
\cmidrule(lr){2-3}\cmidrule(lr){4-5}\cmidrule(lr){6-7}
\textbf{Model} & W & C & W & C & W & C \\
\midrule
Whisper-Turbo  & \textbf{14.76} & 3.48 & 39.56 & 13.70 & \textbf{48.35} & 12.90 \\
IndicWav2Vec   & 14.89 & \textbf{3.07} & 40.68 & 12.74 & 51.65 & 15.38 \\
Whisper-Medium & 15.57 & 3.85 & \textbf{39.06} & 12.92 & 48.98 & \textbf{12.74} \\
Conformer-Hi   & 26.28 & 6.61 & 41.05 & 13.84 & 58.49 & 17.06 \\
XLSR-53        & 26.85 & 5.99 & 57.78 & 14.81 & 62.78 & 16.50 \\
MMS-1B         & 27.28 & 6.06 & 39.83 & \textbf{10.66} & 58.65 & 15.08 \\
\bottomrule
\end{tabular}}
\caption{Fine-tuned benchmark WER/CER (\%) on three test sets.}
\label{tab:benchmark}
\end{table}

\subsection{Inference efficiency}
\label{sec:efficiency}
Real-Time Factor was measured on a single NVIDIA L4 GPU with batch size 1, simulating single-utterance real-time inference. All six models operate well below the real-time threshold. IndicWav2Vec averages RTF $\approx$0.0026, approximately 400$\times$ real-time; Conformer-Hi is marginally faster owing to its smaller parameter count. CTC decoding is structurally cheaper than autoregressive decoding because it requires a single encoder forward pass rather than per-token generation. The practical consequence: Whisper-Turbo, despite matching IndicWav2Vec on accuracy, is approximately 29$\times$ slower (0.076 average RTF vs.\ 0.0026), which is decisive for any deployment with strict latency budgets.

\begin{table}[t]
\centering
\small
\begin{tabular}{lccc}
\toprule
\textbf{Model} & \textbf{OpenSLR} & \textbf{FLEURS} & \textbf{C.~Voice} \\
\midrule
Conformer-Hi   & \textbf{0.0020} & \textbf{0.0019} & \textbf{0.0017} \\
IndicWav2Vec   & 0.0025 & 0.0030 & 0.0024 \\
XLSR-53        & 0.0080 & 0.0088 & 0.0074 \\
MMS-1B         & 0.0214 & 0.0230 & 0.0197 \\
Whisper-Medium & 0.0850 & 0.0826 & 0.0890 \\
Whisper-Turbo  & 0.0979 & 0.0460 & 0.0832 \\
\bottomrule
\end{tabular}
\caption{Real-Time Factor on three test sets (lower is faster).}
\label{tab:rtf}
\end{table}

\subsection{Generalization}
The WER increase from in-domain (OpenSLR) to out-of-domain test sets exposes the generalization gap directly. MMS-1B shows the smallest FLEURS gap (+12.55~pp), roughly half of the next-best model. The plausible explanation: MMS-1B's $>$1{,}100-language pretraining exposes it to acoustic conditions broader than any single in-domain corpus, so the in-domain to out-of-domain shift is closer to in-distribution from its perspective. Whisper-Turbo is the most consistent performer across all three test sets among the high-accuracy models. Common Voice with its crowd-sourced microphone variability is hard for every model.

\begin{table}[t]
\centering
\small
\begin{tabular}{lcc}
\toprule
\textbf{Model} & \textbf{$\Delta$FLEURS} & \textbf{$\Delta$C.~Voice} \\
\midrule
MMS-1B         & \textbf{+12.55} & \textbf{+31.37} \\
Conformer-Hi   & +14.77 & +32.21 \\
Whisper-Medium & +23.49 & +33.41 \\
Whisper-Turbo  & +24.80 & +33.59 \\
IndicWav2Vec   & +25.79 & +36.76 \\
XLSR-53        & +30.93 & +35.93 \\
\bottomrule
\end{tabular}
\caption{WER increase (percentage points) from OpenSLR.}
\label{tab:gen}
\end{table}

\subsection{Training dynamics}
\begin{figure}[t]
\centering
\includegraphics[width=\columnwidth]{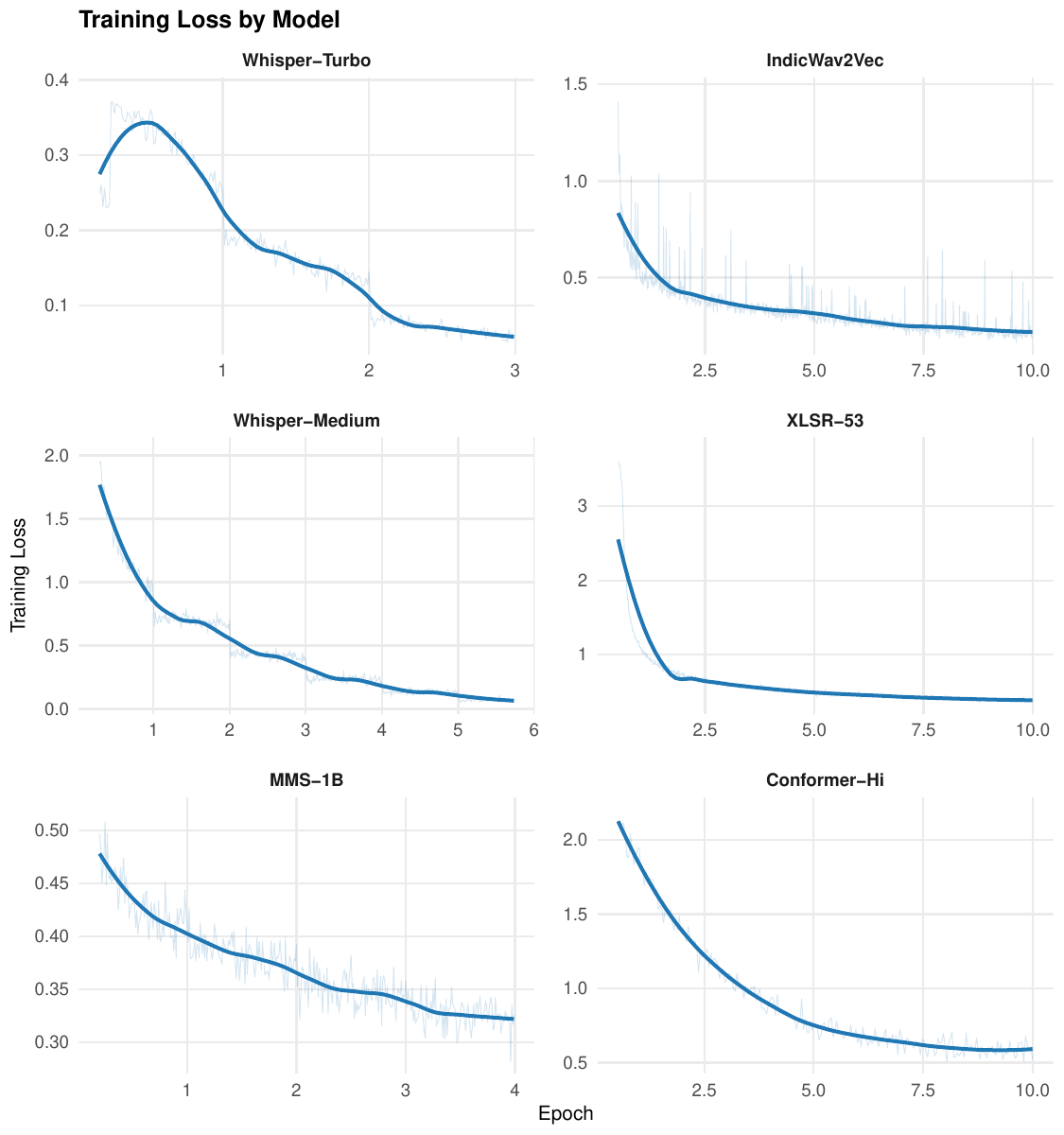}
\caption{Per-model training-loss curves. Y-axis is free per panel; X-axis adapts to each model's actual training horizon.}
\label{fig:loss}
\end{figure}

\begin{figure}[t]
\centering
\includegraphics[width=\columnwidth]{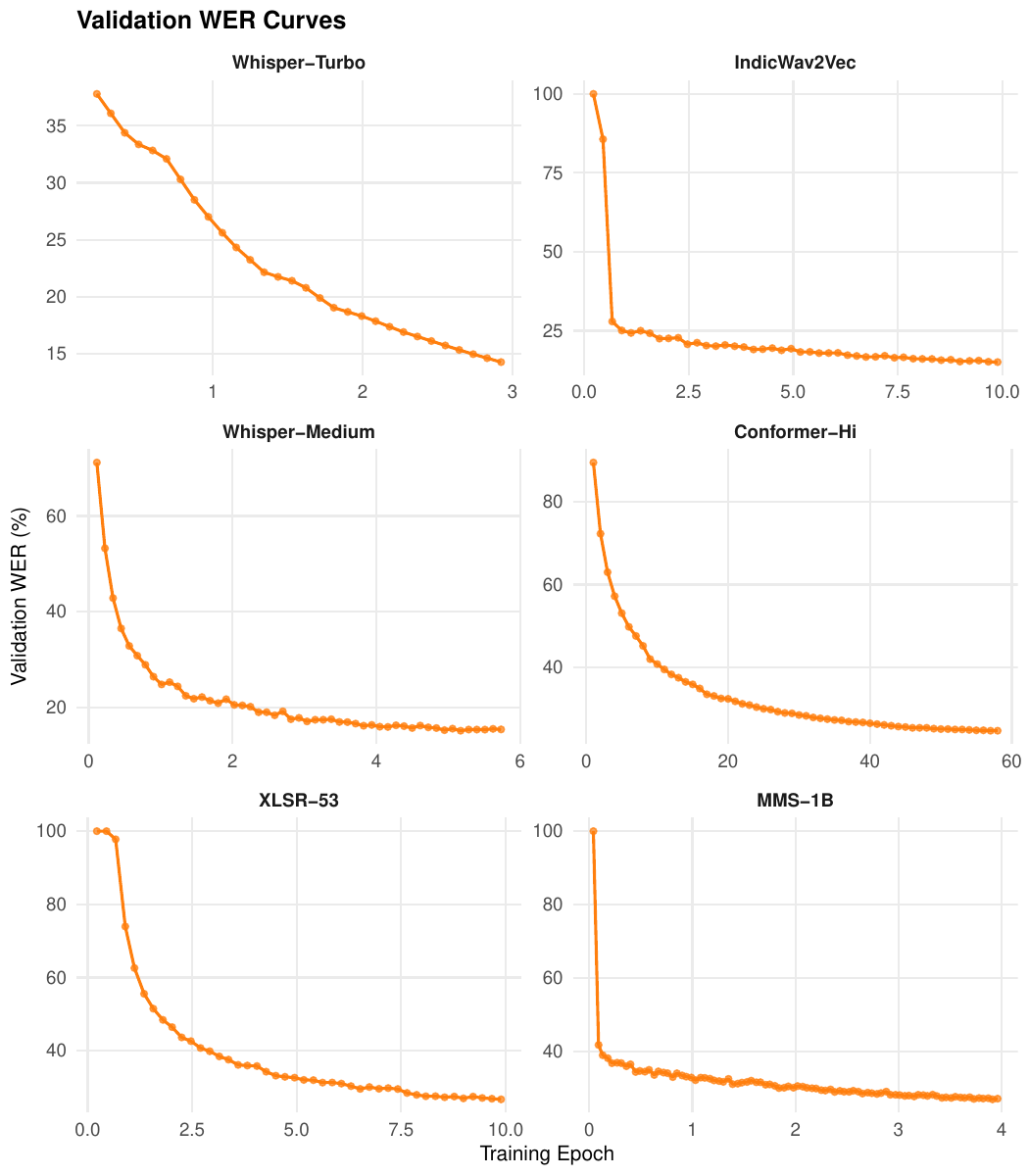}
\caption{Per-model validation-WER trajectories ordered best $\to$ worst: Whisper-Turbo $\to$ IndicWav2Vec $\to$ Whisper-Medium $\to$ Conformer-Hi $\to$ XLSR-53 $\to$ MMS-1B.}
\label{fig:valwer}
\end{figure}

Figure~\ref{fig:loss} shows per-model training-loss curves with x-axis adapted to each model's actual training horizon. Most models converge within the first one to three epochs; early stopping triggered on Whisper-Turbo (epoch 3), Whisper-Medium (epoch 6), and MMS-1B (epoch 4). Figure~\ref{fig:valwer} reports the corresponding validation-WER trajectories, with panels ordered best to worst final WER. The smooth monotone descent on every panel indicates no training instability and no overfitting within the controlled budget.

\subsection{Visual comparison}
\begin{figure}[t]
\centering
\includegraphics[width=\columnwidth]{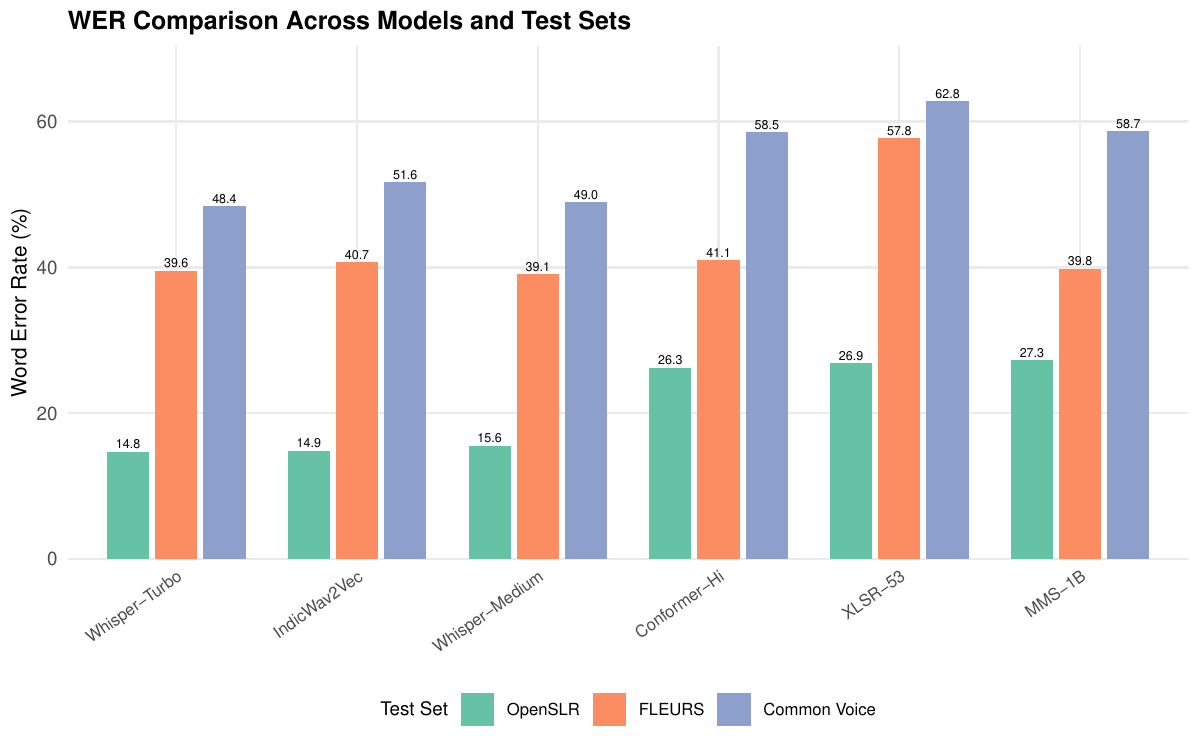}
\caption{Fine-tuned WER across all six models and three test sets.}
\label{fig:werbar}
\end{figure}

\begin{figure}[t]
\centering
\includegraphics[width=\columnwidth]{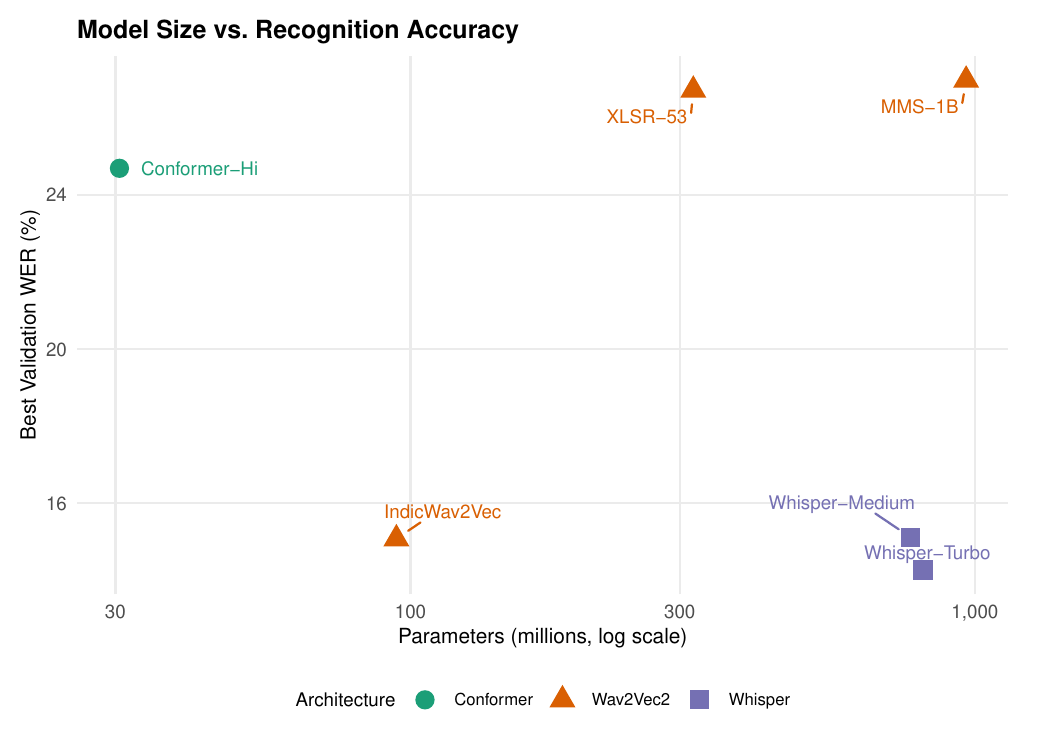}
\caption{Model size (log) versus best validation WER. Smallest two language-family-proximate models reach competitive or better WER than several broadly multilingual models 3--30$\times$ their size.}
\label{fig:sizewer}
\end{figure}

\begin{figure}[t]
\centering
\includegraphics[width=\columnwidth]{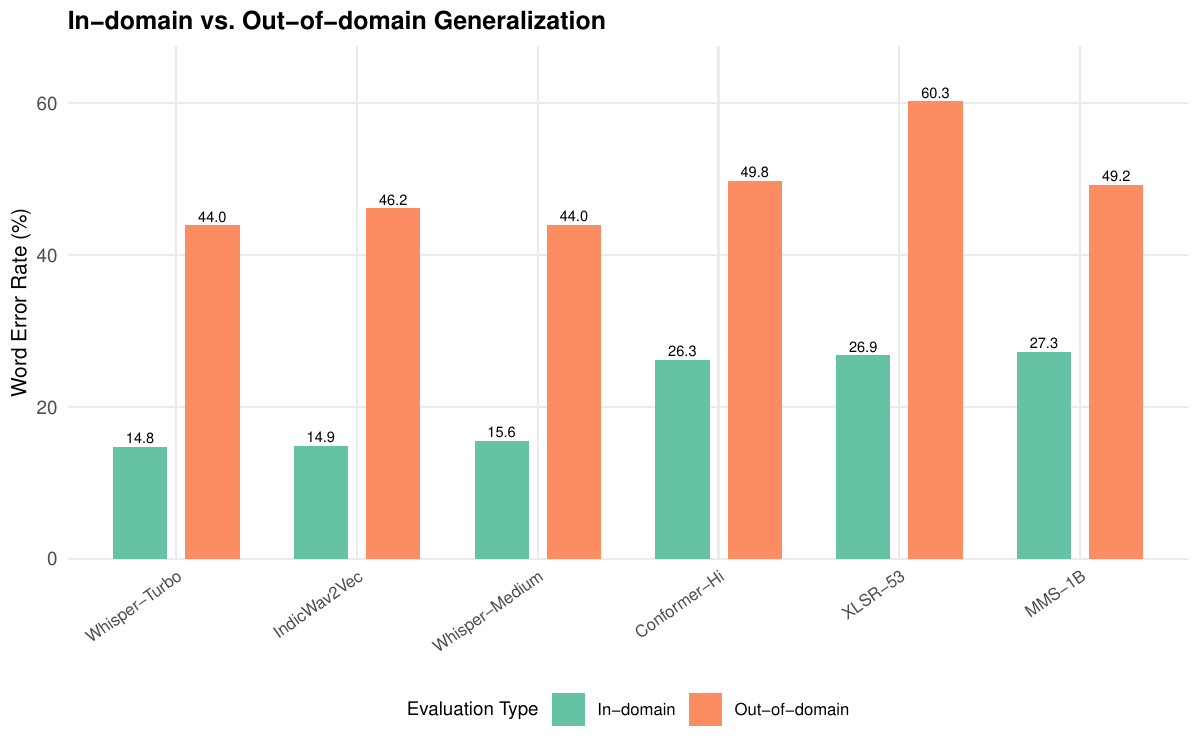}
\caption{In-domain (OpenSLR) versus out-of-domain (mean of FLEURS and Common Voice) WER per model.}
\label{fig:domain}
\end{figure}

\begin{figure}[t]
\centering
\includegraphics[width=\columnwidth]{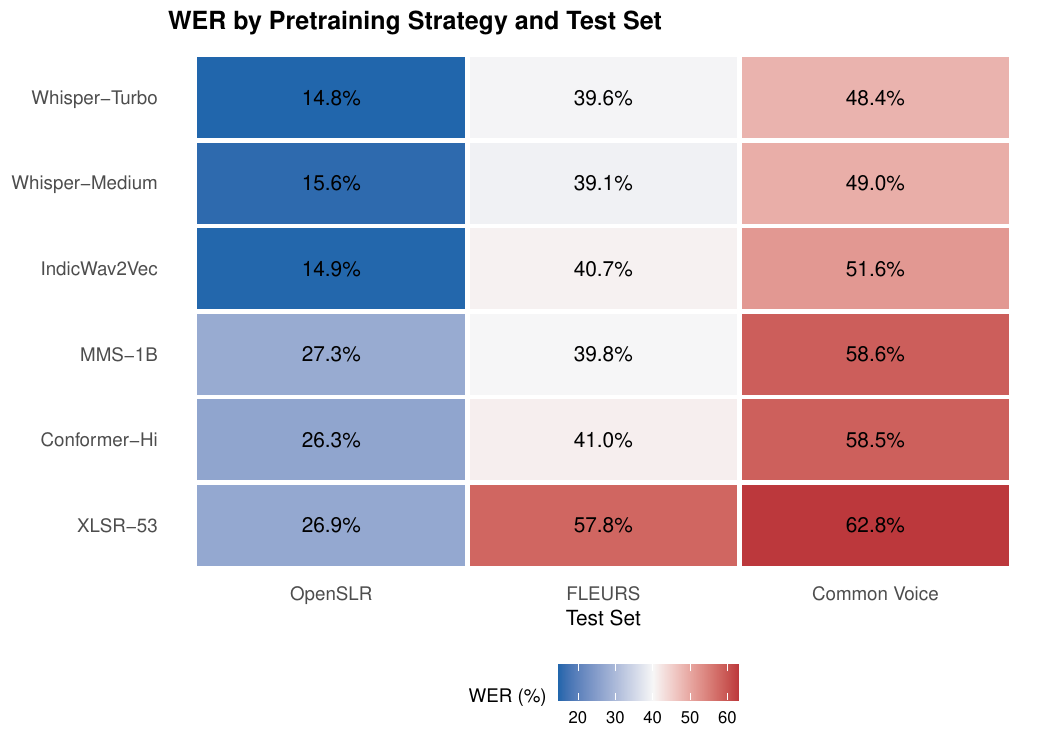}
\caption{WER heatmap organised by pretraining strategy and test set. Indic-proximate and supervised multilingual rows dominate in-domain; the MMS row shows the most uniform colour band, evidencing robustness across domains.}
\label{fig:heat}
\end{figure}

A clustered-bar view of WER across the three test sets, a size-vs-WER scatter, an in-domain versus out-of-domain summary, and a pretraining-strategy heatmap are provided as Figures~\ref{fig:werbar}--\ref{fig:heat}. The smallest two pretraining-proximate models (IndicWav2Vec at 94.4~M, Conformer-Hi at 30.5~M) reach competitive or better WER than several broadly multilingual models that are 3--30$\times$ larger, making the proximity-over-scale finding visible at a glance.

\section{Discussion}
\label{sec:discussion}

\subsection{Proximity vs.\ scale}
The dominant comparative pattern is that linguistic proximity in pretraining outweighs raw scale for in-domain Nepali. IndicWav2Vec (94.4~M parameters, 17~k~hr of Indic-family pretraining) matches Whisper-Turbo (809~M parameters, 680~k~hr of supervised multilingual pretraining) on OpenSLR WER. Conformer-Hi (30.5~M, Hindi-only) beats XLSR-53 (317~M, 53 languages) and MMS-1B (965~M, $>$1{,}100 languages). Going from 53 to $>$1{,}100 pretraining languages does not improve in-domain WER. Within the Whisper family, scaling from Medium to Turbo yields only $\sim$0.4~pp improvement, indicating diminishing returns beyond $\sim$769~M parameters.

\subsection{CTC vs.\ autoregressive trade-off}
At equivalent accuracy, CTC decoding is decisively faster than autoregressive decoding. Whisper-Turbo and IndicWav2Vec tie on OpenSLR WER, but Whisper-Turbo is $\sim$29$\times$ slower. This flips the practical deployment preference toward CTC whenever a latency budget exists. Autoregressive decoders retain advantages for tasks requiring language identification, timestamped transcription, or translation, but for pure ASR with strict latency targets, CTC wins.

\subsection{Scale buys robustness, not accuracy}
MMS-1B is only mid-tier on in-domain accuracy but has the smallest in-domain to out-of-domain WER gap. The plausible mechanism is exposure to broad acoustic conditions during pretraining, which compresses the distribution shift between curated read speech and other domains. Massively multilingual pretraining therefore plays a different role than Indic-specific pretraining: it buys generalization rather than peak accuracy.

\subsection{Comparison with prior published Nepali results}
\citet{ghimire2023active} reported MMS-1B Nepali CER of 6.77\% with active-learning-based data selection on an in-house Nepali set. The MMS-1B CER of 6.06\% reported here on the OpenSLR test partition is lower despite no active-learning intervention, suggesting that the controlled fine-tuning protocol used in this study is at least competitive with active-learning data selection. \citet{pratap2023scaling} and \citet{javed2021towards} report only aggregated Indic numbers; the per-language Nepali measurements here fill that gap.

\subsection{Practical recommendations}
Three deployment-time recommendations follow directly from the measurements. \textbf{IndicWav2Vec} is the preferred choice for real-time and edge use where its 94.4~M parameter count and fast CTC decoder are decisive. \textbf{Whisper-Turbo} is preferred when cross-domain robustness on cleanly recorded speech matters more than latency. \textbf{MMS-1B} is preferred when out-of-domain generalization, rather than peak in-domain accuracy, is the priority. All six fine-tuned checkpoints are released on the Hugging Face Hub for direct download or programmatic loading.\footnote{Models: \href{https://huggingface.co/sumanpaudel1997}{huggingface.co/sumanpaudel1997}.} Integrating a Nepali language model during decoding (shallow fusion for CTC, $n$-best rescoring for Whisper) is the lowest-effort follow-up step expected to yield further WER gains on noisy speech.

\section{Limitations}

\textbf{Compute.} Experiments were carried out on a single NVIDIA L4 GPU (24~GB VRAM), with select models replicated on an A100 80~GB instance for time-bound runs. Hardware constraints limited per-device batch sizes and, for the largest models, the number of training epochs. \textbf{Pipeline.} Conformer-Hi required a separate training pipeline (NVIDIA NeMo) which, while configured to match the shared preprocessing protocol, is not bit-for-bit identical to the Hugging Face fine-tuning pipeline used for the other five models. \textbf{Speech style.} All evaluation data is read speech from curated corpora; performance on spontaneous conversational speech, code-switched (Nepali--English/Hindi) speech, and dialectal variation remains outside this study's scope. \textbf{Decoding.} No external language model was used during decoding, so reported numbers reflect purely acoustic-model performance and may understate what shallow-fusion or rescoring approaches could achieve. \textbf{Pretraining.} Publicly released pretrained checkpoints are used as-is; continued self-supervised pretraining on unlabelled Nepali audio was not attempted and could plausibly close the residual gap to the strongest models.

\section{Conclusion}

We presented the first controlled multi-model multi-dataset benchmark for Nepali ASR. Six pretrained models from three architectural families (CTC self-supervised, autoregressive encoder--decoder, and hybrid Conformer-CTC) were fine-tuned on OpenSLR~SLR54 under an identical protocol and evaluated on OpenSLR, FLEURS, and Common Voice across WER, CER, and RTF. The two top models (Whisper-Turbo, IndicWav2Vec) tie within 0.13~pp despite a 9$\times$ parameter gap, providing direct evidence that language-family proximity in pretraining can substitute for raw scale on in-domain Nepali. CTC decoding is up to 29$\times$ faster than autoregressive Whisper at equivalent accuracy, flipping the practical deployment preference toward CTC under any latency budget. Massively multilingual pretraining (MMS-1B) buys out-of-domain robustness rather than peak accuracy. The resulting benchmark and per-scenario recommendations supply the empirically grounded reference numbers that have been missing from the Nepali ASR literature.

\bibliography{references}
% Note: \bibliographystyle{acl_natbib} is already set inside acl.sty (line 195);
% repeating it here triggers BibTeX's "Illegal, another \bibstyle command" warning.

\clearpage
\appendix

\section{Per-Family Training Hyperparameters}
\label{app:hparams}

The complete per-family training hyperparameters used in the controlled fine-tuning protocol are listed below. The Wav2Vec~2.0 family freezes the convolutional feature encoder and updates only the Transformer layers and the CTC head. The Whisper family fine-tunes all parameters at a low learning rate to avoid degrading pretrained representations; Whisper-Large-v3-Turbo additionally required gradient accumulation and approximately 34 hours of training on a single NVIDIA L4 GPU. Conformer-Hi was fine-tuned inside NVIDIA NeMo with cosine annealing. SpecAugment was applied to every model.

\begin{table}[h]
\centering
\footnotesize
\setlength{\tabcolsep}{4pt}
\resizebox{\columnwidth}{!}{%
\begin{tabular}{lccccc}
\toprule
\textbf{Family} & \textbf{LR} & \textbf{Batch} & \textbf{Epochs} & \textbf{Warmup} & \textbf{Schedule} \\
\midrule
Wav2Vec 2.0   & $3\!\times\!10^{-4}$ & 8  & 10 & 500 steps & Cosine decay     \\
Whisper       & $1\!\times\!10^{-5}$ & 8  & 3--6 & Linear  & Linear decay     \\
Conformer-Hi  & $1\!\times\!10^{-4}$ & 16 & 10 & 500 steps & Cosine annealing \\
\bottomrule
\end{tabular}}
\caption{Per-family training hyperparameters. All families used AdamW.}
\label{tab:hparams}
\end{table}

\section{Sample Transcriptions}
\label{app:samples}

Three reference--prediction pairs from the released benchmark, one
from each test set, illustrate qualitative output behaviour. Devanagari is
rendered with the Kohinoor Devanagari font.

\textbf{Example~1: IndicWav2Vec on OpenSLR (clean studio recording).}\\[2pt]
\noindent{\small Reference:\par}
\noindent\hspace*{1em}{\nepalifont नेपालको संविधानले सबै नागरिकलाई समान अधिकार प्रदान गर्दछ।}\par\vspace{2pt}
\noindent{\small Prediction:\par}
\noindent\hspace*{1em}{\nepalifont नेपालको संविधानले सबै नागरिकलाई समान अधिकार प्रदान गर्दछ।}\par\vspace{2pt}
\noindent{\small WER 0.00, CER 0.00.}

\vspace{6pt}
\textbf{Example~2: XLSR-53 on FLEURS (cross-dataset read speech).}\\[2pt]
\noindent{\small Reference:\par}
\noindent\hspace*{1em}{\nepalifont खानेपानीको समस्या धेरै ठाउँमा छ।}\par\vspace{2pt}
\noindent{\small Prediction:\par}
\noindent\hspace*{1em}{\nepalifont कानेपानीको समस्या देरै टाउँमा छ।}

\vspace{6pt}
\textbf{Example~3: MMS-1B on Common Voice (crowd-sourced noisy).}\\[2pt]
\noindent{\small Reference:\par}
\noindent\hspace*{1em}{\nepalifont विद्यार्थीहरूले परीक्षामा राम्रो गरे।}\par\vspace{2pt}
\noindent{\small Prediction:\par}
\noindent\hspace*{1em}{\nepalifont विदयारथीहरूले परीकषामा रामरो गरे।}

\vspace{6pt}
Per-utterance reference/prediction pairs for every (model, test set) combination are available in the released benchmark dataset.\footnote{Dataset: \url{https://huggingface.co/datasets/sumanpaudel1997/nepali-asr-benchmark}}

\section{Reproducibility Notes}
\label{app:repro}

\textbf{Hardware.} Single NVIDIA L4 GPU (24~GB VRAM) for most runs; a single NVIDIA A100 80~GB instance was used for the largest checkpoints when the L4 was VRAM-bound. Whisper-Large-v3-Turbo required gradient accumulation. \textbf{Software.} Hugging Face Transformers and Datasets for the Wav2Vec~2.0 and Whisper families; NVIDIA NeMo for Conformer-Hi. WER and CER were computed with \texttt{jiwer} after NFC normalisation of both reference and hypothesis. RTF was measured with batch size 1 to simulate single-utterance real-time inference. \textbf{Splits.} The 80/10/10 OpenSLR~SLR54 splits enforce speaker disjointness across training, validation, and test partitions. FLEURS and Common Voice were used with their predefined test splits. \textbf{Seeds and determinism.} Random seeds were fixed; mixed-precision (FP16) was used to reduce memory consumption.

\end{document}